\documentclass[letterpaper]{article} % DO NOT CHANGE THIS
\usepackage[preprint]{aaai2027}  % DO NOT CHANGE THIS
\usepackage[hyphens]{url}  % DO NOT CHANGE THIS
\usepackage{graphicx} % DO NOT CHANGE THIS
\usepackage{natbib}  % DO NOT CHANGE THIS AND DO NOT ADD ANY OPTIONS TO IT
\usepackage{caption} % DO NOT CHANGE THIS AND DO NOT ADD ANY OPTIONS TO IT
\usepackage{booktabs}
\usepackage{multirow}
\usepackage{amsmath,amssymb}

\newcommand{\duno}{\mbox{DU-NO}}

\newcommand{\R}{\mathbb{R}}

\title{DU-NO: A Parameter-Efficient Double U-Shaped Neural Operator for Phase-Resolving Wave Modeling}
\author{
    Enrique Hernandez Noguera\textsuperscript{\rm 1},
    Md Meftahul Ferdaus\textsuperscript{\rm 1},
    Nathan Cooper\textsuperscript{\rm 1},\\
    Elias Ioup\textsuperscript{\rm 2},
    Mahdi Abdelguerfi\textsuperscript{\rm 1}
}
\affiliations{
    \textsuperscript{\rm 1}Canizaro-Livingston Gulf States Center for Environmental Informatics,\\
    LSU New Orleans, New Orleans, LA, USA\\
    \textsuperscript{\rm 2}U.S. Naval Research Laboratory, Stennis Space Center, MS, USA
}

\begin{document}

\maketitle

\begin{abstract}
Phase-resolving wave models such as FUNWAVE-TVD are the accuracy
standard for nearshore dynamics, resolving the shoaling, refraction, and
breaking of individual waves, but their cost rules them out for the
ensembles, uncertainty quantification, and real-time warning that
operational forecasting demands. Neural operators promise solver-level
accuracy at a fraction of that cost, yet on wave-dominated fields the
accurate ones are large: hybrid spectral--convolutional operators such
as U-FNO (the strongest baseline in our study after \duno{}) buy their
fidelity with tens of millions of parameters. We introduce \duno{}
(\emph{Double U-shaped Neural Operator}), a multiscale U-shaped spectral
operator that attaches lightweight convolutional U-Net branches only at
its two shallowest encoder and decoder levels. The placement follows a
sampling argument: high-wavenumber content exists only on fine grids, so
the local, full-band pathways go where that content lives, while the
coarse, band-limited levels stay purely spectral. A depth-decaying mode
schedule holds the model to 3.64M~parameters, an order of magnitude
below U-FNO. On our publicly released FUNWAVE-TVD benchmark, \duno{}
attains the best autoregressive rollout error of six identically trained
architectures, improving on U-FNO by $14.9\%$ with $10.8\times$ fewer
parameters, and a frequency-band analysis shows the gain holds across
\emph{all} bands, including the high-wavenumber band where
truncated-spectral operators collapse. Parameter-matched controls
confirm the gain is architectural: rescaled to the same $3.6$M budget,
the best baseline still trails \duno{} by $28.6\%$. The advantage carries beyond
nearshore waves: \duno{} matches the strongest baselines on 2D
Navier--Stokes and wins clearly on PDEBench shallow-water rollouts.
Code, trained models, and evaluation artifacts are available at
\url{https://anonymous.4open.science/r/duno-code-5A7B/}.
\end{abstract}

\section{Introduction}
\label{sec:intro}

% P1 — the application gap: phase-resolving wave models are accurate but
% prohibitively expensive; surrogates promise ensemble/real-time use.
Coastal flooding and wave-driven overtopping of seawalls and dunes are
among the most damaging coastal hazards, and the same nearshore dynamics
govern beach erosion, navigation safety, and coastal-defense design.
Anticipating them requires resolving how waves shoal, refract, and break
as they approach shore, the capability that separates phase-resolving models
from the phase-averaged spectral models used for basin-scale forecasting
\citep{ferdaus2025wavereview}. Phase-resolving Boussinesq models such as FUNWAVE-TVD
\citep{shi2012funwave} are built for precisely this and remain the
accuracy standard for nearshore dynamics. However, that fidelity carries a steep
computational cost that makes the solver too slow for many operational
uses. The remedy is a surrogate that reproduces the solver's phase-resolved
output in a fraction of the time. Neural operators \citep{kovachki2023neural}, which learn
resolution-robust mappings between function spaces, have emerged as
surrogates that are orders of magnitude faster than the solvers that
generated their training data, with the Fourier neural operator
\citep[FNO;][]{li2021fourier} and its descendants
\citep{wen2022ufno,rahman2023uno,tran2023factorized} being the dominant
family for regular-grid problems.

% P2 — the gap and the design principle (architecture-led, sampling
% insight; the theory is made precise in the method).
On wave-dominated dynamics the most accurate operators are hybrids that
pair a global spectral pathway with local convolution, because the steep
fronts, breaking, and dispersive tails that carry the physics live at high
wavenumbers that a truncated spectral parameterization cannot represent.
The leading hybrid, U-FNO~\citep{wen2022ufno}, secures that fidelity by
running its spectral stack at full resolution and adding full-resolution
convolutional U-Net branches, at a cost of tens of millions of parameters.
We observe that this full-resolution cost is avoidable: most spectral work
captures large-scale structure, which a multiscale backbone represents far
more cheaply by coarsening progressively. Coarsening also changes where
local convolution helps: each step
strips the highest-wavenumber content a convolutional branch is meant to
recover, so that content, and the branch's value, survives only on the finest
grids. A sampling argument (Section~\ref{sec:method-branches}) makes this
concrete, and our ablations confirm it. Placing compact U-Net branches in parallel
with the spectral path only at those shallow levels is what makes the design
parameter-efficient, matching or improving a full-resolution hybrid's accuracy
at a fraction of its parameters.

% P3 — DU-NO concretely.
We turn this principle into \duno{} (\emph{Double U-shaped Neural
Operator}), a multiscale, U-shaped spectral operator augmented with
compact convolutional U-Net branches
\citep{ronneberger2015unet,wen2022ufno} that run \emph{in parallel} with
the spectral path, but \emph{only at the two shallowest encoder and
decoder levels}. Because the retained-mode budget shrinks and the channel
width is capped as the pyramid coarsens, the model is small,
3.64M~parameters against U-FNO's 39.4M; placement
ablations confirm the principle, with U-Net branches at the bottleneck actively
hurting accuracy at higher cost, because the high-wavenumber content such a
branch would recover has already been stripped away by the downsampling that
precedes that level.

% P4 — the evidence: fair protocol, three datasets, headline numbers.
We evaluate six architectures
(\duno{}, U-FNO~\citep{wen2022ufno}, U-NO~\citep{rahman2023uno},
Transolver~\citep{wu2024transolver}, U-Net~\citep{ronneberger2015unet}, and
FNO~\citep{li2021fourier}), each trained under one
identical configuration for three seeds. Our primary benchmark, and the
application this work targets, is a dataset of phase-resolving nearshore
wave fields that we generate with the FUNWAVE-TVD solver
\citep{shi2012funwave}. To probe generalization beyond it, we add two
standard operator-learning benchmarks, 2D Navier--Stokes
\citep{li2021fourier} and the PDEBench shallow-water equations
\citep{takamoto2022pdebench}. \duno{} attains the best
autoregressive rollout accuracy of the suite on the FUNWAVE waves it
targets, improving over the strongest competitor, U-FNO, by $14.9\%$ in
rel-$L_2$ (relative $L_2$ error, the field-reconstruction error normalized
by the field's own magnitude; lower=better) while using $10.8\times$ fewer parameters at comparable
per-step latency; on the Navier-Stokes benchmark it stays competitive with the best baselines,
while surpassing them with a clear gap on the shallow-water equations. A frequency band-resolved analysis (Section~\ref{sec:results}) bears out
the mechanism behind this sampling argument: truncated spectral operators
concentrate their error in the high-wavenumber band that carries most of
the energy of wave fields, while \duno{} stays accurate across the
spectrum.

% P5 — contributions: architecture (with theory) + empirical.
\textbf{Contributions.}
\begin{itemize}
  \item \textbf{A scale--frequency placement principle.} We give a
  sampling-based rule for \emph{where} local convolutional capacity belongs
  in a multiscale spectral operator: a convolutional branch can only help in
  the per-level \emph{spectral gap} between the retained Fourier modes and
  the grid's Nyquist limit (the highest wavenumber the grid can represent),
  the band the truncated spectral path discards
  (Section~\ref{sec:method-branches}). Because each coarsening step strips
  high-wavenumber content, that gap, and with it a branch's value, falls
  steadily with depth, from $68\%$ of field energy at the finest level to
  $17\%$ at the coarsest (measured on held-out fields in
  Section~\ref{sec:method-branches}; see Appendix~E.1 in the supplementary
  material). Because
  it depends only on a level's sampling structure, not on \duno{} specifics,
  the rule sets branch placement \emph{before} training; a placement
  ablation confirms its prediction.
  \item \textbf{\duno{}.} We realize the
  principle as a double U-shaped operator (Section~\ref{sec:method}): a
  U-NO-style spectral pyramid carrying U-FNO-style convolutional branches at
  \emph{only} its two shallowest levels, with a depth-decaying mode schedule
  and capped channel growth that track each level's shrinking Nyquist band.
  This holds the model to 3.64M~parameters, and ablations corroborate both
  the placement and the mode schedule.
  \item \textbf{A phase-resolving nearshore-wave benchmark.} We build and
  publicly release (access in the supplementary material) a controlled $3^4$
  factorial FUNWAVE-TVD dataset
  (Section~\ref{sec:setup}) that stresses the steep, breaking, high-wavenumber
  regime underrepresented in the smooth, diffusion-dominated problems common
  in operator-learning evaluation.
  \item \textbf{Fair, mechanism-level empirical validation.} Under one
  identical protocol across six architectures and three seeds, \duno{} is the
  most accurate model on the FUNWAVE-TVD waves benchmark, improving on the strongest
  baseline, U-FNO, by $14.9\%$ at $10.8\times$ fewer parameters, $4.4\times$
  less memory, and comparable latency; parameter-matched controls, with
  every baseline rescaled to \duno{}'s budget under the same protocol,
  separate this gain from raw capacity effects; applied unchanged to 2D
  Navier--Stokes and the shallow-water equations it stays competitive (tied
  on Navier--Stokes, best long-horizon on shallow water). A reusable band-resolved error analysis localizes
  each family's failure to a specific wavenumber range and shows \duno{}'s
  gain holds across \emph{all} bands (Section~\ref{sec:results}).
\end{itemize}

\section{Related work}
\label{sec:related}

\paragraph{Neural operators for PDE surrogacy.}
Operator learning replaces a numerical solver with a learned map between
function spaces \citep{kovachki2023neural,lu2021deeponet}. On regular
grids, the Fourier neural operator \citep{li2021fourier} parameterizes a
global kernel by a truncated set of Fourier modes and underpins most
state-of-the-art surrogates. A long line of successors restructures the
spectral core for capacity or efficiency: F-FNO factorizes the spectral
weights across dimensions \citep{tran2023factorized}, tensorized variants
compress them \citep{kossaifi2023multigrid}, multiwavelet operators
change the basis \citep{gupta2021multiwavelet}, convolutional neural
operators replace the spectral product with alias-aware convolutions
\citep{raonic2023cno}, and recent work drops the kernel's structural
assumptions altogether by learning its singular-value decomposition
\citep{koren2026svdno}. A parallel thread learns attention over latent
physics tokens \citep{wu2024transolver}, recently recast as linear
attention for efficiency \citep{hu2026linearno}. Our contribution is
orthogonal to all of these: we keep the vanilla spectral block and change
\emph{where} a complementary convolutional pathway is placed within a
multiscale operator.

\paragraph{U-shaped and hybrid spectral--convolutional operators.}
Two prior lines combine U-Net structure with spectral operators, and
\duno{} draws on both. U-FNO \citep{wen2022ufno} runs a single-scale
full-resolution spectral stack with a parallel U-Net in its later
``U-Fourier'' layers, whose branches hold much of its $\sim$39.4M
parameters. U-NO \citep{rahman2023uno} arranges spectral blocks in a
multiscale encoder--decoder but stays purely spectral. \duno{} keeps the
strengths and sheds the costs: a U-NO-style spectral pyramid with
U-FNO-style branches \emph{only at the two shallowest levels}, tying
placement to the pyramid's scale--frequency structure via a sampling
argument (Section~\ref{sec:method-branches}) and a placement ablation
(Section~\ref{sec:ablations}), giving 3.64M parameters against U-FNO's
39.4M at better accuracy.

\paragraph{Spectral bias and high-frequency error in learned surrogates.}
Neural networks fit low frequencies first \citep{rahaman2019spectral};
spectral truncation in FNOs hard-codes an additional architectural ceiling
above the retained modes. Over-smoothing of fine scales is a
recognized failure of spectral surrogates in turbulence and weather
emulation \citep{pathak2022fourcastnet}, and the error such surrogates
accumulate over long autoregressive rollouts can compound until the
prediction leaves the physical solution manifold entirely
\citep{huang2026physicscorrect}. We contribute a controlled,
cross-architecture quantification of \emph{where} in wavenumber each family's
error concentrates, and connect that high-frequency fidelity to long-horizon
rollout stability (Section~\ref{sec:results}; Appendix~F in the
supplementary material).

\paragraph{Machine learning for ocean-wave forecasting.}
A separate line of work applies learning directly to wave forecasting
rather than to solver surrogacy, and targets integrated or point-wise
quantities: physics-informed networks forecast buoy time series
\citep{schmidt2024pinn}, learned models coupled with data assimilation
correct surface-wave forecasts \citep{pokhrel2024assimilation}, losses
tailored to differential structure sharpen such predictions
\citep{schmidt2026ratio}, and learned classifiers flag extreme events
including rogue waves \citep{pokhrel2020rogue}. These operate on
phase-averaged or station-level data and do not reconstruct the wave field
itself. \duno{} addresses the complementary problem: reproducing a
phase-resolving solver's full spatial field autoregressively at
solver-level fidelity.

\paragraph{Wave dynamics as an operator-learning benchmark.}
Operator-learning evaluations have centered on a small set of canonical
problems, many of them smooth and diffusion-governed, such as Darcy flow,
whose solutions are spectrally compact and lose little when their upper
modes are truncated. Phase-resolving \emph{nearshore} wave dynamics, where
steep, breaking, and dispersive fields concentrate their energy at high
wavenumbers and a low-mode spectral parameterization is most exposed,
remain comparatively underrepresented. We adopt this setting as our
primary benchmark. Nearshore wave fields from FUNWAVE-TVD
\citep{shi2012funwave}, a community-standard phase-resolving Boussinesq
solver occupying the highest-fidelity tier of the coastal wave-model
hierarchy \citep{ferdaus2025wavereview}, give a demanding and
application-relevant testbed for operator learning. To establish that the resulting model is not narrowly tuned to
our proposed dataset, we additionally evaluate on two standard
operator-learning benchmarks from the same broad family of fluid dynamics:
the PDEBench shallow-water equations \citep{takamoto2022pdebench}, a
depth-averaged free-surface system in the lineage of the Boussinesq
equations FUNWAVE solves, and the canonical 2D Navier--Stokes problem
\citep{li2021fourier}, the smoother, spectrally compact regime the FNO
family already handles well.

\section{The Double U-shaped Neural Operator (\duno{})}
\label{sec:method}

\begin{figure*}[tp]
  \centering
  \includegraphics[width=0.9\textwidth]{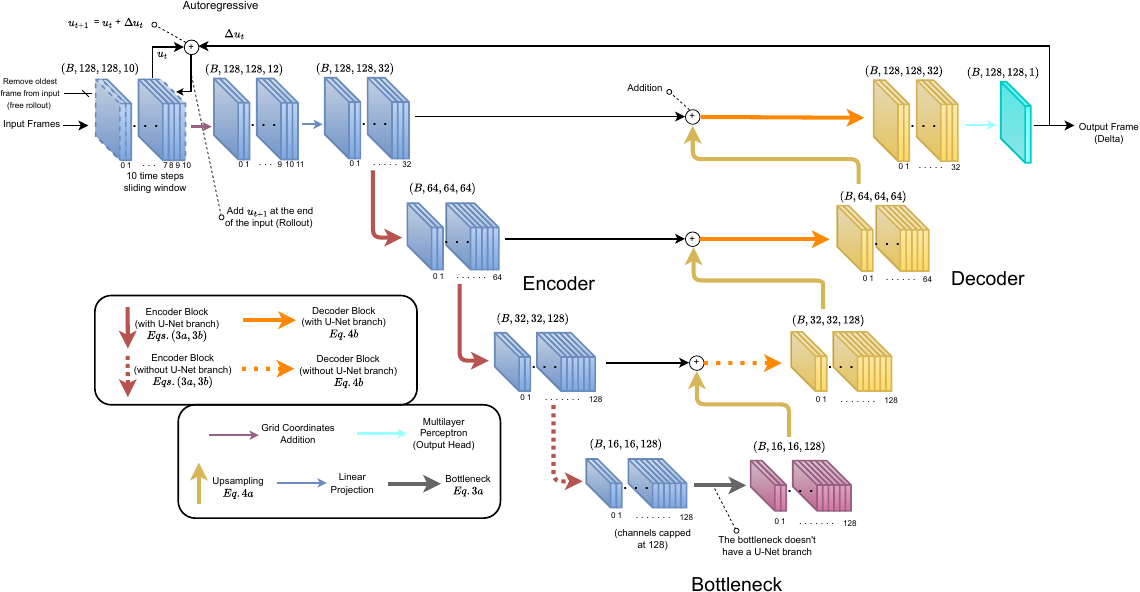}
  \caption{\duno{}. The 10-frame input window is lifted by a linear
  projection with two coordinate channels, passed through a three-level
  encoder that halves resolution and grows width ($32{\to}64{\to}128$), a
  bottleneck at the coarsest $16\times16$ scale (with channels capped at 128), and a three-level decoder
  that upsamples and fuses the encoder's skip connections; an MLP head emits the increment
  added to the last frame. Convolutional U-Net branches run in parallel with
  the spectral path only at the two shallowest encoder and decoder levels
  (Section~\ref{sec:method-branches}).}
  \label{fig:arch}
\end{figure*}

\duno{} is a multiscale spectral operator whose two shallowest encoder
and decoder levels carry parallel convolutional U-Net branches
(Figure~\ref{fig:arch}). We describe in turn the autoregressive problem
it solves, the spectral backbone and its depth-decaying capacity
schedule, the convolutional branches and the scale--frequency argument
that motivates their placement, and the resulting parameter and latency cost.

\subsection{Problem setup}
\label{sec:method-setup}

% Local leading bump: the tall inline math on the "consecutive solution
% frames" and "$G_\theta:...$" lines otherwise nearly touch; raising
% \lineskip(limit) adds ~2pt clearance only where lines run tight, leaving
% ordinary lines untouched. Scoped to this paragraph.
{\lineskip=2pt\lineskiplimit=2pt%
We learn an autoregressive surrogate for 2D time-dependent dynamics. Given
a window of $T_{\mathrm{in}}$ consecutive solution frames
$u_{t-T_{\mathrm{in}}+1:t} \in \R^{H\times W\times T_{\mathrm{in}}}$, the
model $G_\theta\!:\R^{H\times W\times T_{\mathrm{in}}}\!\to\R^{H\times W}$
predicts the \emph{increment} to the \emph{single} next frame,
\begin{equation}
  \hat{u}_{t+1} \;=\; u_t + G_\theta\!\left(u_{t-T_{\mathrm{in}}+1:t}\right),
  \label{eq:residual}
\end{equation}
and longer trajectories come only from re-applying this one-frame map,
sliding the window over its own predictions (free rollout). Residual prediction
(Eq.~\ref{eq:residual}) makes the learning target the \emph{change} of
the field rather than its magnitude; it is a standard, consistency-motivated ingredient in
autoregressive neural PDE solvers \citep{brandstetter2022mppde} that all six
models use, so it is controlled across the comparison rather than part of
\duno{}'s contribution. Training is
teacher-forced: the one-frame map is applied $T_{\mathrm{out}}$ times per
window to predict $T_{\mathrm{out}}$ successive increments, each step
conditioned on ground-truth history (not the model's own prediction), and
the loss is the sum of per-step relative $L_2$ errors on the increments.
Evaluation is always free rollout (Section~\ref{sec:setup}). We use
$T_{\mathrm{in}} = T_{\mathrm{out}} = 10$ throughout, except where we study
an extended rollout.\par}

\subsection{A multiscale spectral backbone with depth-decaying mode budget}
\label{sec:method-backbone}

\paragraph{Spectral blocks.}
The backbone is built from FNO-style spectral convolutions
\citep{li2021fourier}: for a feature map $v$ with $C$ channels,
\begin{equation}
  \mathrm{Spectral}(v) \;=\; \mathcal{F}^{-1}\!\big(R \cdot
  \mathbf{1}_{|k| < M_\ell}\,\mathcal{F}(v)\big),
  \label{eq:spec}
\end{equation}
where $\mathcal{F}$ is the 2D real FFT (fast Fourier transform), which maps
the feature map to complex Fourier coefficients indexed by an integer
wavenumber $k = (k_x, k_y)$. The indicator $\mathbf{1}_{|k| < M_\ell}$ indicates that only
low wavenumbers (modes) are kept, with $|k| := \max(|k_x|, |k_y|)$ and $M_\ell$ the per-level
retained-mode budget set by the depth-decaying schedule below. 
$R$ then applies a learned complex $C\times C$ matrix to each retained mode. Each
spectral convolution is followed by a pointwise two-layer MLP, and runs
in parallel with a $1\times1$ convolution that acts as a full-band linear
path.

\paragraph{Encoder--decoder pyramid.}
\duno{} arranges these blocks in a U-shape (Figure~\ref{fig:arch}): a
three-level encoder that halves the spatial resolution (while doubling the channels until reaching the capped amount) 
at each step, a bottleneck at the coarsest scale, and a three-level decoder that restores
the original resolution. Following standard FNO practice
\citep{li2021fourier}, the $T_{\mathrm{in}}$ input frames are first
augmented with two coordinate channels holding each grid point's
normalized position in $[0,1]^2$; a linear layer applied at each grid point
then lifts these $T_{\mathrm{in}}{+}2$ values to a feature map $v_1$ with
$c_1 = 32$ channels. Encoder level $\ell$ then computes
\begin{subequations}\label{eq:encoder}
\begin{gather}
  s_\ell = \sigma\big(\mathrm{MLP}(\mathrm{Spectral}_\ell(v_\ell)) + W_\ell v_\ell
            \,[+\,\mathrm{UNet}_\ell(v_\ell)]\big), \label{eq:encoder-s}\\
  v_{\ell+1} = \sigma\big(P_\ell\,(\mathrm{AvgPool}_{2\times2}(s_\ell))\big),
            \label{eq:encoder-v}
\end{gather}
\end{subequations}
with $\sigma$ the GELU activation. The first line forms the level feature
$s_\ell$ as the sum of three complementary paths: the spectral path of
Eq.~\ref{eq:spec} with its pointwise MLP (global mixing, restricted to the
retained modes), the full-band pointwise residual $W_\ell$ (a $1\times1$
convolution), and, only at the shallow levels, the
convolutional U-Net branch supplying the local full-band mixing the
truncated spectral path lacks. The second line prepares the next, coarser
level: $2\times2$ average pooling halves the resolution and the
$1\times1$ projection $P_\ell$ grows the channel count. Each $s_\ell$ is
also passed directly to the matching decoder level as a skip connection,
preserving detail that pooling discards. Decoder level $\ell$ mirrors
this,
\begin{subequations}\label{eq:decoder}
\begin{gather}
  m_\ell = \sigma\big(\sigma(Q_\ell\,(\mathrm{Up}(z_{\ell+1}))) + s_\ell\big), \label{eq:decoder-m}\\
  z_\ell = \sigma\big(\mathrm{MLP}(\mathrm{Spectral}_\ell(m_\ell)) + W_\ell m_\ell
            \,[+\,\mathrm{UNet}_\ell(m_\ell)]\big),
            \label{eq:decoder-z}
\end{gather}
\end{subequations}
where the feature $z_{\ell+1}$ from the coarser level below is bilinearly
upsampled ($\mathrm{Up}$), projected to the skip's width by the
$1\times1$ map $Q_\ell$ and activated, fused with the skip $s_\ell$ by
addition under a second activation, and processed by the same three-path
block. The bottleneck is a single multipath block (Eq.~\ref{eq:encoder-s}) at the coarsest scale that
carries no convolutional branch (Section~\ref{sec:method-branches}); a
pointwise MLP head maps the final decoder feature to the increment of
Eq.~\ref{eq:residual}.

\paragraph{Mode and width schedules.}
Capacity follows the pyramid. Retained modes within the Fourier operator
halve with depth, starting at 12 and clipping at 2, giving $12/6/3$
across encoder and decoder levels, and $2$ at the bottleneck. The modes
thus halve in lockstep with each level's \emph{Nyquist band}, the range of
wavenumbers its grid can represent (bounded above by the Nyquist limit, the
highest wavenumber resolvable at a given sampling density), which halves
each time average pooling halves the resolution. Every level therefore
retains the same fraction of its representable spectrum (up to the clip); fine
structure above the retained band is left to the shallow levels, their
skips, and the branches of
Section~\ref{sec:method-branches}. Channels move the opposite way,
doubling with depth but capped at four times the base width
($32/64/128$ for encoder and decoder levels, and $128$ at the bottleneck). The two schedules balance cost: the
spectral weights $R$ scale as $C^2 M_\ell^2$ (a $C\times C$ matrix per
retained mode), so doubling $C$ while halving $M_\ell$ leaves per-level
spectral cost roughly constant, and the cap keeps the nearly mode-free
coarsest levels from hoarding parameters. As a result, the full model has
3.64M parameters. We validate the mode schedule empirically in
Section~\ref{sec:ablations} (Table~\ref{tab:ablations}).

\subsection{Selectively-placed Convolutional U-Net branches}
\label{sec:method-branches}

The two paths described so far mix information in complementary but
incomplete ways: the spectral path is global yet band-limited to its
$M_\ell$ retained modes, and the $1\times1$ path is full-band yet cannot
mix neighboring grid points. What neither provides is \emph{local,
full-band} spatial mixing, which is exactly what a convolutional network
offers. \duno{}'s defining choice is therefore a third path, a compact
three-level convolutional U-Net \citep{ronneberger2015unet} built from
$3\times3$ convolutions and adopted from U-FNO's U-Fourier block
\citep{wen2022ufno}, run in parallel with the other two paths and summed
with them before the activation (Eq.~\ref{eq:encoder-s}, bracket). The branch
module is identical to U-FNO's U-Fourier-block U-Net; what \duno{} contributes
is not the module but \emph{where} it runs. Since
every branch adds parameters and latency, the question the design must
answer is \emph{where} in the pyramid such a branch earns its cost.

\paragraph{Where can a convolutional branch help?}
The answer follows from how the pyramid distributes frequency content.
Encoder level $\ell$ acts on a feature map downsampled by $2^{\ell-1}$
through average pooling, which low-pass filters and subsamples. By the
Nyquist--Shannon sampling theorem \citep{shannon1949}, the grid at level
$\ell$ can represent wavenumbers only up to a Nyquist limit
$k^{\mathrm{Nyq}}_\ell = k^{\mathrm{Nyq}}_1 / 2^{\ell-1}$, halving with
depth; at each pooling step the averaging filter strongly attenuates the
upper half-band, and what survives subsampling folds onto lower
wavenumbers. Either way, high-wavenumber structure (steep fronts,
breaking, dispersive tails) exists only on the finest grids.

\begin{figure}[tbp]
  \centering
  \includegraphics[width=0.9\linewidth]{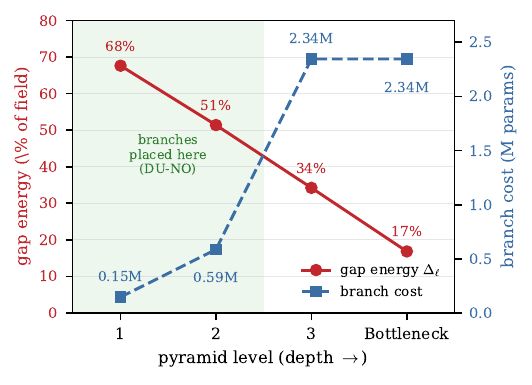}
  \caption{Spectral-gap energy vs.\ convolutional-branch cost across pyramid
  levels. Gap energy $\Delta_\ell$ (Eq.~\ref{eq:gap}, red) falls with depth
  while a U-Net branch's parameter cost (blue) rises; they cross at the edge
  of the shallow region where \duno{} places its branches, so branch leverage
  (energy per parameter) collapses from the finest level
  (level~1) to the bottleneck.}
  \label{fig:gap}
\end{figure}

Within a level, the spectral block acts only on the $M_\ell$ retained
modes and zeroes the \emph{spectral gap}
$M_\ell \le |k| \le k^{\mathrm{Nyq}}_\ell$. A $3\times3$ convolution is
the complementary primitive: local in space, its frequency response has
no structural zeros, so a stack of such kernels can read from and write
to the entire gap the spectral path discards; inside the retained band it
is largely redundant, since the spectral block already applies a learned
matrix to every mode there. The leverage a branch can add at level $\ell$
is therefore limited by the signal that occupies the gap,
\begin{equation}
  \Delta_\ell \;=\; \int_{M_\ell \le |k| \,\le\, k^{\mathrm{Nyq}}_\ell}
  S_\ell(k)\,\mathrm{d}^2k,
  \label{eq:gap}
\end{equation}
with $S_\ell$ the power spectrum of the field at level $\ell$'s resolution:
for a truth frame $u$, $S_\ell(k) = \mathbb{E}\big[\,|\mathcal{F}(u^{(\ell)})(k)|^2\,\big]$,
where $u^{(\ell)}$ is $u$ after $\ell{-}1$ rounds of the $2\times2$ average
pooling of Eq.~\ref{eq:encoder-v} ($u^{(1)}=u$, the full-resolution field) and
$\mathbb{E}$ averages over the test frames. Evaluating
Eq.~\ref{eq:gap} on $3{,}840$ held-out test frames as the gap-band share of
field power (Parseval-normalized), the gap holds $68\%$ of the
field's energy at the finest level, falling to $51\%$, $34\%$, and $17\%$
at the coarser scales, while a branch's cost grows $16\times$ over the
same span (it scales with the squared channel width); Figure~\ref{fig:gap}
plots this measured gap-energy profile against the branch cost.
The same profile governs the decoder, whose level $\ell$ operates on the
same grid with the skip connection re-injecting the full level-$\ell$ band.

\paragraph{The resulting placement, and its validation.}
The efficient allocation is thus to spend convolutional capacity where
$\Delta_\ell$ is large and leave the coarse, already band-limited levels
purely spectral. \duno{} accordingly places branches only at the
\emph{two shallowest encoder and decoder levels}; the sampling argument identifies these
levels as where branches should help, and we test that prediction directly with a
placement ablation (Table~\ref{tab:ablations},
Section~\ref{sec:ablations}), which confirms it at both extremes:
shallow-only placement matches or improves on branches at every level at
under half the parameters, removing the shallow branches degrades
accuracy, and branches at the bottleneck, where the gap energy is
smallest, \emph{hurt} accuracy at higher cost. We also test whether simply
enlarging the branches helps, doubling their channel width at the same
placement (Table~\ref{tab:ablations}): this keeps error within seed noise, while
inflating the branch parameters to $8.08$M ($2.2\times$).
Added branch capacity therefore buys no real accuracy, so the
canonical \duno{} keeps standard-width branches.

\subsection{Cost analysis}
\label{sec:method-cost}

\duno{}'s efficiency relative to U-FNO comes from two compounding choices.
The backbone is multiscale, so most blocks run at reduced resolution rather
than U-FNO's full-resolution spectral stack and wide full-resolution U-Net
branches; and the depth-decaying modes ($M_\ell = 12/6/3/2$) with channels
capped at $128$ keep both the spectral weights and the two branch pairs
compact. The result is 3.64M parameters against U-FNO's 39.4M ($10.8\times$
fewer), $4.4\times$ lower peak inference memory (57 vs.\ 252\,MB at
batch~1), and comparable latency (3.3 vs.\ 4.1\,ms), at better accuracy
(Table~\ref{tab:main}).

\paragraph{Latency and ensemble throughput.}
The parameter gap does not yield a proportional latency gap: at batch~1 per-step
cost is dominated by kernel-launch and memory overhead on an under-utilized GPU,
so a $10.8\times$ smaller model runs at comparable, not $10.8\times$ lower,
latency. These cross-model GPU latencies are measured on a consumer
RTX~4080~SUPER (Table~\ref{tab:main}); the same comparison on a datacenter
NVIDIA~H100~NVL, where batch-1 under-utilization is larger still, is reported in
the supplementary material (Appendix~E.4). The footprint instead buys
throughput, a single $16$\,GB GPU advancing an eight-member ensemble in one pass
at $4.7\times$ lower per-member cost (supplementary material, Appendix~E.3), the
axis large operational ensembles need. The decisive gain, though, is over the
solver \duno{} replaces: on the same $16$-core Ryzen~9~7950X3D CPU that runs
FUNWAVE-TVD, \duno{} advances a nearshore forecast a median $173\times$ faster
than the solver (Table~\ref{tab:main}, last column; median over the
$81$-simulation benchmark, a like-for-like CPU-to-CPU comparison).
% NOTE(2026-07-15): the CL-CPU-24 run IS DU-NO's own CPU inference, not the
% solver's --- artifact records tool="surrogate", params_M=3.6389, and it was
% written by time_duno.py, which only times DU-NO. "24" was torch's thread
% count (torch.get_num_threads()), NOT a core count.
% NOTE(resolved): 5.4/5.7 ms latency = main-table cross-model pair on the eval
% GPU (same machine, DU-NO vs U-FNO); 4.7x = within-card batch-1->8 ratio on the
% 4080 SUPER. Each is internally same-machine; the text states the 4.7x as a
% within-card ratio and never combines it with the absolute ms numbers.

\section{Experimental setup}
\label{sec:setup}

\paragraph{Benchmarks.}
Our generated FUNWAVE-TVD nearshore-wave dataset is the primary benchmark;
Navier--Stokes and shallow water are standard fluid dynamics operator-learning benchmarks
we add to confirm the model generalizes beyond the main benchmark.
We evaluate on three 2D time-dependent problems:
\begin{itemize}
  \item \textbf{FUNWAVE-TVD nearshore waves} (primary; generated by us):
  phase-resolving surface-elevation ($\eta$) fields from the fully
  nonlinear, dispersive FUNWAVE-TVD Boussinesq solver \citep{shi2012funwave}
  with shock-capturing wave breaking, run over the measured bathymetry of
  the U.S.\ Army Corps of Engineers Field Research Facility at Duck, North
  Carolina \citep{birkemeier2000frf,collins2018frf}. We build it as a controlled $3{\times}3{\times}3{\times}3$
  factorial study of $81$ simulations (57/8/16 train/val/test)
  over the four dominant forcing parameters (significant wave height, spectral peak frequency, principal
  direction, and wavemaker depth), spanning calm to energetic, short to long
  period, and obliquely to normally incident seas; each contributes $240$
  frames cropped to the active surf zone and interpolated to
  $128\times128$. Full solver configuration, the regime grid, and the windowing procedure are
  given in Appendix~B of the supplementary material.
  \item \textbf{Navier--Stokes}: the canonical FNO 2D vorticity
  benchmark at the \emph{turbulent} viscosity $10^{-5}$
  \citep{li2021fourier}, where a forward enstrophy cascade fills the
  field with fine vortical filaments; 1{,}200 trajectories of 20 steps at
  $64\times64$, split 840/120/240 by trajectory (one window per
  trajectory).
  \item \textbf{Shallow water}: the PDEBench 2D radial dam-break
  \citep{takamoto2022pdebench}, a hyperbolic problem whose fine scales
  are the sharp propagating bore fronts; 1{,}000 trajectories of 100 steps at
  $128\times128$, split 700/100/200 by trajectory (only the first
  10-in/10-out window per trajectory is used for training; rationale in
  Appendix~B).
\end{itemize}
All splits are deterministic, trajectory-disjoint, and shared by every
model. On every benchmark a model takes a 10-frame input window and predicts
autoregressively, one frame at a time: we evaluate a 10-step rollout for
FUNWAVE and Navier--Stokes, and, because shallow water's 10-step dam-break
transient is degenerate (all models hit a floor near $10^{-4}$ rel-$L_2$), a
90-step rollout is used (Table~\ref{tab:generalization}), which spans essentially
the entire trajectory.

\paragraph{Baselines.}
U-FNO, U-NO, Transolver, FNO, and a convolutional
U-Net make up our baselines. U-Net and FNO
double as an ablative decomposition of \duno{} into its convolutional and
spectral constituents; U-FNO is the strongest baseline and the closest
hybrid; U-NO shares \duno{}'s U-shaped multiscale spectral backbone; and
Transolver is included as a competitive transformer-based operator.
Baseline configurations follow their reference implementations. In
addition, every baseline is retrained at a \emph{parameter-matched}
configuration, rescaled to \duno{}'s $3.6$M budget by a single capacity
knob with the protocol otherwise unchanged; the exact configurations and
resulting parameter counts are listed in Appendix~C.1 of the
supplementary material.

\paragraph{Training protocol.}
Comparisons of operator architectures are notoriously sensitive to
per-model tuning. We therefore freeze a single training protocol per
dataset and apply it to all six architectures: AdamW \citep{loshchilov2019adamw}
($\mathrm{lr}=10^{-3}$, weight decay $10^{-5}$), OneCycle schedule
\citep{smith2019super} ($\mathrm{max\_lr}=10^{-3}$, 30\% warm-up,
cosine), 250 epochs, relative-$L_2$ loss on residual targets,
teacher-forced 10-step training, batch size 8, gradient clipping at norm
1.0. No data augmentation, auxiliary losses, or compression is applied to any model.

\paragraph{Seeds, selection, and reporting.}
Each (model, dataset) pair on the three main benchmarks is trained with
three seeds controlling initialization (i.e., minibatch order; splits are
fixed independently of the training seed). Model selection uses
validation rollout error; all test numbers are computed at the
best-validation checkpoint (and report mean~$\pm$~std over seeds). The
ablation sweep of Table~\ref{tab:ablations} is single-seed exploratory
work, reported as such.

\paragraph{Metrics.}
Following PDEBench \citep{takamoto2022pdebench} and standard neural-operator
practice, our primary metric is free-rollout relative $L_2$ over the 10-step
horizon. Alongside it, Table~\ref{tab:main} reports the parameter count,
batch-1 peak inference memory, and batch-1 latency,
and we decompose the rollout error by radial wavenumber
$k=\sqrt{k_x^2+k_y^2}$ (distinct from the model's max-norm $|k|$) into the
low ($k\in[0,4]$), mid ($k\in[5,12]$), and high ($k\ge13$) bands of PDEBench's
fRMSE metric (Figure~\ref{fig:bands}). Additional metrics are defined and reported in Appendix~D of the supplementary material.

\section{Results}
\label{sec:results}

% ---------------------------------------------------------------------
% R1. Main comparison table — wired to make_paper_tables.py output.
% Regenerate with:  python make_paper_tables.py [--project ...]
% then copy/symlink the emitted .tex into paper/tables/.
% ---------------------------------------------------------------------
\begin{table}[tp]
  \centering
  \caption{Main comparison on our FUNWAVE-TVD benchmark: free-rollout
  rel-$L_2$, parameters, batch-1 peak memory, batch-1 GPU (consumer
  RTX~4080~SUPER) and CPU ($16$ threads, matching the solver's $16$ MPI ranks
  on the same processor) latency, and end-to-end speedup over FUNWAVE-TVD for the same
  forecast window (median over the $81$-simulation benchmark). Metrics are
  mean$\pm$std over 3 seeds (peak memory and latency are constant within
  seeds); best value bold.
  $^\dagger$Single seed under the identical protocol, at the architecture's
  published reference configuration; GPU memory and latency reported where
  measured (``--'' otherwise).
  $^\ddagger$Single-seed control with the architecture rescaled to
  \duno{}'s $3.6$M parameter budget under the identical protocol.}
  \label{tab:main}
  \resizebox{\columnwidth}{!}{% Seeded P0 results (6 models x 3 seeds). rel-L2 and fRMSE_hi are the
\setlength{\tabcolsep}{4pt}
\begin{tabular}{lcccccc}
\toprule
Model & rel-$L_2$ & Params & Mem (MB) & GPU ms & CPU ms & Speedup \\
\midrule
KNO$^\dagger$   & $0.4389$ & 0.26M & 27 & 1.3 & 24.6 & 151 \\
F-FNO$^\dagger$ & $0.3868$ & 0.16M & 23 & 1.9 & 17.0 & 219 \\
FNO            & $0.3206 \pm 0.0076$ & 1.21M  & 94 & 1.0 & 27.2 & 137 \\
U-Net          & $0.2019 \pm 0.0152$ & 31.0M  & 212 & 1.8 & 37.0 & 100 \\
Transolver     & $0.1588 \pm 0.0117$ & 11.2M  & 189 & 25.0 & 527.5 & 7 \\
CNO$^\dagger$   & $0.1531$ & 10.6M & 59 & 4.3 & 101.5 & 37 \\
U-NO           & $0.1522 \pm 0.0085$ & 107.9M & 953 & 5.6 & 195.2 & 19 \\
MWT$^\dagger$   & $0.1290$ & 1.67M & 104 & 12.6 & 53.9 & 69 \\
U-FNO          & $0.1111 \pm 0.0034$ & 39.4M  & 252 & 4.1 & 59.8 & 62 \\
\duno{} (ours) & $\mathbf{0.0946 \pm 0.0020}$ & 3.64M & 57 & 3.3 & 21.5 & 173 \\
\midrule
\multicolumn{7}{l}{\emph{Baselines rescaled to \duno{}'s parameter budget}} \\
% USER DECISION 2026-07-18: ALL matched controls in the MAIN table ("don't
% leave any baseline's matched size comparison behind"). Filled rows sorted
% worst->best (as the main block); "--" rel-L2 rows are still training or
% awaiting their official eval (job numbers in the header comment).
ONO$^\ddagger$        & $0.5660$ & 3.86M & -- & -- & -- & -- \\
LinearNO$^\ddagger$   & $0.5441$ & 3.72M & -- & -- & -- & -- \\
KNO$^\ddagger$        & $0.5336$ & 3.70M & -- & -- & -- & -- \\
FNO$^\ddagger$        & $0.2937$ & 3.68M & -- & -- & -- & -- \\
U-Net$^\ddagger$      & $0.2470$ & 3.67M & -- & -- & -- & -- \\
F-FNO$^\ddagger$      & $0.2316$ & 3.68M & -- & -- & -- & -- \\
U-NO$^\ddagger$       & $0.1809$ & 3.80M & -- & -- & -- & -- \\
MG-TFNO$^\ddagger$    & $0.1753$ & 3.87M & -- & -- & -- & -- \\
CNO$^\ddagger$        & $0.1658$ & 3.74M & -- & -- & -- & -- \\
Transolver$^\ddagger$ & $0.1637$ & 3.55M & -- & -- & -- & -- \\
MWT$^\ddagger$        & $0.1380$ & 3.92M & -- & -- & -- & -- \\
LSM$^\ddagger$        & $0.1232$ & 3.69M & -- & -- & -- & -- \\
U-FNO$^\ddagger$      & $0.1217$ & 3.85M & -- & -- & -- & -- \\
\bottomrule
\end{tabular}
}
\end{table}

\paragraph{Main comparison on FUNWAVE (Table~\ref{tab:main}).}
\duno{} attains the best rollout accuracy of the suite,
$0.0946 \pm 0.0020$ rel-$L_2$ against $0.1111 \pm 0.0034$ for U-FNO (the second best operator tested), a
$14.9\%$ improvement that holds on every one of the $16$ held-out test simulations
(paired $p = 3.1\times10^{-5}$; supplementary material, Appendix~E.5); it does so with
$10.8\times$ fewer parameters, $4.4\times$ lower peak inference memory than
U-FNO, and comparable batch-1 latency.
The remaining models trail by ${\sim}60$--$70\%$ (U-NO, Transolver), rising
to more than $3\times$ (FNO). Four further operators, each run once under the
same frozen protocol at its published reference configuration, span a wide
range: MWT is strongest at $0.1290$ ($36\%$ above \duno{}, ahead of the far
larger U-NO and Transolver), followed by CNO ($0.1531$), while F-FNO and KNO
trail far behind ($0.39$, $0.44$).

% Matched-controls paragraph. Numbers = OFFICIAL paper_eval rollout metric
% (2026-07-19; ALL 13 landed + typeset 2026-07-22). Best control U-FNO 0.1217
% (nearest below it is LSM 0.1232, still >0.1217), so "None comes close" holds.
\paragraph{Parameter-matched controls (Table~\ref{tab:main},
$^\ddagger$).}
Holding capacity fixed does not close the gap. Rescaled to \duno{}'s
$3.6$M budget under the identical protocol, the oversized models get
worse (U-FNO $0.1111\to0.1217$, Transolver $0.1588\to0.1637$, U-NO
$0.1522\to0.1809$, U-Net $0.2019\to0.2470$) and the undersized FNO
improves only modestly ($0.3206\to0.2937$). None comes close to \duno{}:
the best control, U-FNO at $0.1217$, still trails it by $28.6\%$, so the
advantage in the reference-size comparison cannot be attributed to
capacity allocation. The matched configurations and sizing methodology
are given in Appendix~C.1 of the supplementary material.

% ---------------------------------------------------------------------
% R2. Frequency-space analysis = empirical confirmation of the sampling
% argument (figure moved here from the intro) + qualitative rollouts.
% Figure built by figures/make_motivation_figure.py from
% architectures/derisk/derisk_out/raw.npz; refresh on seeded ckpts.
% ---------------------------------------------------------------------
\begin{figure}[tbp]
  \centering
  \includegraphics[width=0.9\linewidth]{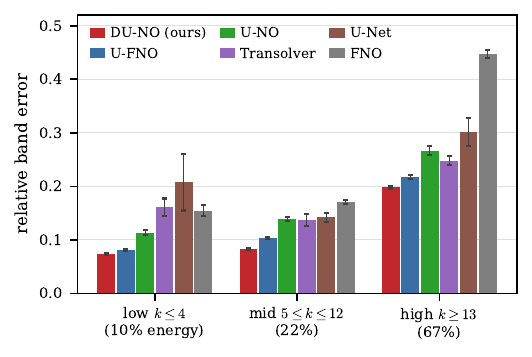}
  \caption{Frequency-band error of FUNWAVE free rollouts (192 test windows;
  mean$\pm$std over 3 seeds), each band's error relative to that band's truth
  RMS. \duno{} is lowest in every band; the pure-spectral FNO is worst in the high
  band.}
  \label{fig:bands}
\end{figure}

\paragraph{Frequency-space behavior confirms the sampling argument
(Figure~\ref{fig:bands}).}
Decomposing free-rollout error by wavenumber band 
exposes how each family fails, as Section~\ref{sec:method-branches} predicts.
The pure FNO concentrates error in the high band ($k\ge13$), by far the largest
high-band error of the suite; U-NO, also purely spectral, shares that high-band weakness but
more mildly, while Transolver is flat but weak at low wavenumbers and U-Net is
weak throughout. The hybrids lead in every band, and
\duno{} is lower than U-FNO in \emph{all three}, at an order of magnitude fewer parameters.

% ---------------------------------------------------------------------
% R3. Robustness: long rollouts, regimes, resolution, domain statistic.
% ---------------------------------------------------------------------
\paragraph{Beyond the training horizon.}
Pushed from 10 rollout steps to 90, the FUNWAVE ranking holds with no
crossover: \duno{} keeps the lowest error at every step and decorrelates
slowest, its spatial correlation staying above $\rho=0.9$ for $42$ steps
against $39$ for U-FNO and ${\le}27$ for the others (FNO first, at step $11$).
Extended-rollout curves are in the supplementary material (Appendix~E.2).

% ---------------------------------------------------------------------
% R4. Ablations.
% ---------------------------------------------------------------------
\subsection{Ablations}
\label{sec:ablations}

\begin{table}[tbp]
  \centering
  \caption{Architecture ablations (FUNWAVE benchmark): rel-$L_2$ as the
  convolutional branches are placed at different pyramid levels. All rows are
  single seed, while the canonical \duno{} represents the 3-seed mean (as a
  reference). enc=encoder, dec=decoder.}
  \label{tab:ablations}
  \small
  \scalebox{0.9}{\begin{tabular}{lcc}
    \toprule
    Configuration & Params & rel-$L_2$ \\
    \midrule
    \duno{} canonical, 3 seeds (Table~\ref{tab:main}) & 3.64M & $0.0946 \pm 0.0020$ \\
    \midrule
    \multicolumn{3}{l}{\emph{Branch placement}} \\
    none (pure spectral pyramid)              & 2.17M  & 0.1794 \\
    shallowest enc+dec level                  & 2.47M  & 0.1076 \\
    two shallowest enc+dec levels (\duno{})   & 3.64M  & 0.0937 \\
    all enc+dec levels                        & 8.33M  & 0.0962 \\
    all levels + bottleneck                   & 10.67M & 0.0996 \\
    deepest enc+dec level                     & 6.86M  & 0.1717 \\
    bottleneck only                           & 4.52M  & 0.1844 \\
    wider branches, \duno{} placement         & 8.08M  & 0.0931 \\
    \midrule
    \multicolumn{3}{l}{\emph{Mode schedule (\duno{} placement; modes per level)}} \\
    $12/12/12/8$ (no decay)                   & 16.22M & 0.0967 \\
    $12/6/3/2$ (\duno{})                      & 3.64M  & 0.0937 \\
    $8/4/2/2$ (aggressive decay)              & 2.66M  & 0.0946 \\
    \bottomrule
  \end{tabular}}
\end{table}

\paragraph{Placement ablation (Table~\ref{tab:ablations}).}
The ablation varies only U-Net branch placement (single seed), holding backbone and protocol
fixed. One branch pair at the shallowest level already recovers most of the
gain over the branch-free pyramid ($0.1076$ rel-$L_2$), and a second pair at
the next level is best ($0.0937$). The same pair placed at the deepest levels
loses most of the benefit ($0.1717$ vs.\ $0.1076$) at nearly $3\times$ the
parameters, and a bottleneck-only branch is worse than none ($0.1844$ vs.\ $0.1794$). More capacity does not help:
all-level placement trails \duno{} at $2.3\times$ the parameters, and doubling
branch width shifts error by less than seed noise. Convolutional capacity
pays off exactly where the sampling argument says high-frequency energy
survives, and nowhere else.

\paragraph{Mode schedule.}
Modes decay with depth ($12/6/3/2$) to track each level's shrinking Nyquist
band; holding them constant ($12/12/12/8$) costs $4.5\times$ the parameters
for no gain (16.22M, 0.0967 vs.\ 0.0937), while pushing further ($8/4/2/2$,
single seed) reaches 2.66M within seed noise of the canonical model
($0.0946$). We keep $12/6/3/2$ because its modes halve with each level's Nyquist 
band rather than being simply tuned for minimal parameters.

% ---------------------------------------------------------------------
% R5. Generalization to other PDE regimes (NS, SWE) — secondary evidence
% that the FUNWAVE-tuned design is not wave-specific. Closes the Results.
% ---------------------------------------------------------------------
\subsection{Generalization beyond FUNWAVE-TVD waves}
\label{sec:generalization}

\begin{table}[tbp]
  \centering
  \caption{Generalization beyond our FUNWAVE benchmark. Free-rollout
  rel-$L_2$ (3 seeds); 10-step horizon for Navier--Stokes, 90-step for
  shallow water under the first-window stress-test protocol (rationale in
  Section~\ref{sec:setup}). The full-trajectory column is the
  conventional-protocol control (single seed, same 90-step evaluation):
  every model but U-Net and Transolver saturates the benchmark below $0.002$,
  so it no longer discriminates (Appendix~E.9 of the supplementary material).
  \duno{} stays competitive on both regimes at far fewer parameters.}
  \label{tab:generalization}
  \scalebox{0.8}{% NS column: filled 2026-06-13 from the executed {model}_ns_s{seed}.ipynb
% notebooks (test rollout rel-L2 at the best-val checkpoint, "Best Test
% L2"), 10-step horizon (NS trajectories are only 20 frames). On NS the
% top is a DU-NO / U-FNO statistical tie, so both are bold.
% SWE first-window column: 90-step long-horizon free rollout (mean +/- std
% over 3 seeds of ext_rel_l2 from
% paper_eval_out/<model>_swe_s{seed}/raw_metrics.npz, via
% paper_aggregate.py). The 10-step radial dam-break transient saturates
% every model near 1e-4 rel-L2 and does not discriminate; the long horizon
% does. Bold = best or within one std of best (DU-NO only). Transolver
% diverges (>1) at long horizon.
% SWE full-trajectory column (added 2026-07-19): the conventional-protocol
% control, single seed, same 90-step eval
% (paper_eval_out/<model>_sweft_s0/raw_metrics.npz, mean ext_rel_l2).
% All completed models but U-Net AND Transolver saturate (<=0.002, no
% decorrelation in 90 steps); U-FNO nominally lowest -> bold. Transolver cell
% FILLED 2026-07-22: 0.1142 (worst full-traj, decorr 46.8). Details: supp E.9.
\begin{tabular}{lccc}
\toprule
& & \multicolumn{2}{c}{Shallow water} \\
\cmidrule(lr){3-4}
Model & Navier--Stokes & first-window & full-traj. \\
\midrule
FNO            & $0.0796 \pm 0.0022$ & $0.0840 \pm 0.0055$ & $0.0019$ \\
U-Net          & $0.3655 \pm 0.0311$ & $0.0688 \pm 0.0114$ & $0.0137$ \\
U-NO           & $0.1667 \pm 0.0031$ & $0.0785 \pm 0.0091$ & $0.0008$ \\
Transolver     & $0.2242 \pm 0.0021$ & $1.9207 \pm 1.8662$ & $0.1142$ \\
U-FNO          & $\mathbf{0.0735 \pm 0.0010}$ & $0.0831 \pm 0.0027$ & $\mathbf{0.0006}$ \\
\duno{} (ours) & $\mathbf{0.0740 \pm 0.0011}$ & $\mathbf{0.0552 \pm 0.0046}$ & $0.0011$ \\
\bottomrule
\end{tabular}
}
\end{table}

\paragraph{The design transfers beyond FUNWAVE-TVD waves
(Table~\ref{tab:generalization}).}
Applied unchanged under the same protocol to the canonical 2D
Navier--Stokes benchmark \citep{li2021fourier} and the PDEBench
shallow-water equations \citep{takamoto2022pdebench}, \duno{} stays among
the top models. On Navier--Stokes, the FNO family's home turf, it ties U-FNO
within seed noise and beats the
rest; on shallow water, whose 10-step transient is degenerate (all models hit a floor
near $10^{-4}$ rel-$L_2$), we compare rollout rel$L_2$ at
90 steps, where \duno{} is most accurate by a clear margin, while Transolver
diverges. The first-window training protocol behind this comparison is a
% Transolver full-traj RESOLVED 2026-07-22: ext90=0.1142, decorr 46.8 (NOT
% saturated -> WORST full-traj model). "all but U-Net" -> "U-Net and
% Transolver" updated here + gen-table caption + E.9 caption + both table cells.
deliberate beyond-horizon stress test: the control in the table's last
column, which instead trains on the full trajectories, drives most
architectures to near-interpolation accuracy (rel-$L_2\le0.002$ with no
decorrelation over the whole 90 steps) for all but U-Net and Transolver, so
the conventionally trained benchmark stops separating them
altogether (details in Appendix~E.9 of the supplementary material).
\duno{}'s performance is therefore not FUNWAVE-specific.

\section{Conclusion}
\label{sec:conclusion}

We presented a scale--frequency placement principle for hybrid multiscale spectral
operators: local convolutional capacity earns its cost only in the spectral
gap a sampling argument locates at the shallowest pyramid levels, and \duno{}
is the parameter-efficient operator that places compact U-Net branches
exactly there. On our
FUNWAVE-TVD benchmark, under one protocol across six architectures and three
seeds, \duno{} is the most accurate of the suite at $10.8\times$ fewer
parameters and $4.4\times$ less memory than U-FNO; it
stays competitive on 2D Navier--Stokes and wins on the shallow-water benchmark
by a clear margin.

\paragraph{Limitations and future work.}
Our study covers 2D problems on regular grids; extending the backbone to
3D or irregular geometries is open. The FUNWAVE surrogate is trained for a
single bathymetry (Duck, NC), so cross-bathymetry generalization, and
transfer from synthetic to real coastlines, remains untested. We also
predict the surface elevation alone, though the solver co-evolves the
horizontal velocities $u,v$, so velocity-dependent quantities such as
nearshore currents are beyond the present model's reach; carrying the full
$(\eta,u,v)$ state is a natural extension. Validating the placement rule
analytically remains open, as does curbing long-rollout error accumulation:
training-free inference-time correction schemes
\citep{huang2026physicscorrect} are orthogonal to the choice of
architecture and could be layered directly on \duno{}, although their
efficient variants assume a linear PDE residual that the Boussinesq
equations do not satisfy.

\bibliography{refs}

% AAAI-27: check whether the reproducibility checklist must be included
% in the paper PDF or is collected via the submission form.
% \input{ReproducibilityChecklist.tex}

\end{document}